\documentclass[lettersize,journal]{IEEEtran}
\usepackage{fancyhdr}
\usepackage{cite}
\usepackage{bm}
\usepackage{booktabs}
\usepackage{hyperref}
\usepackage{amsmath,amssymb,amsfonts}
\usepackage{algorithmicx}
\usepackage{graphicx}
\usepackage{subfigure}
\usepackage{textcomp}
\usepackage{xcolor}
\usepackage{multirow}
\usepackage{CJKutf8}
\usepackage{makecell}
\usepackage{bbding}
\usepackage{pifont}

\newcommand{\argm}[1][x]{\ensuremath{\mathop{\arg\min}\limits_{#1}}}
\newcommand{\fan}[1][x]{\ensuremath{\Vert #1\Vert}}
\def\BibTeX{{\rm B\kern-.05em{\sc i\kern-.025em b}\kern-.08em
    T\kern-.1667em\lower.7ex\hbox{E}\kern-.125emX}}
\usepackage{balance}

\begin{document}
\title{PRISTA-Net : Deep Iterative Shrinkage Thresholding Network for Coded Diffraction Patterns Phase Retrieval}
\author{Aoxu Liu, Xiaohong Fan, Yin Yang, Jianping Zhang
\thanks{This work was supported by the National Natural Science Foundation of China Project (11771369, 12071402), the Education Bureau of Hunan Province, P. R. China (22A0119), the National Key Research and Development Program of China (2020YFA0713503), the Natural Science Foundation of Hunan Province (2020JJ2027, 2023GK2029), and the Postgraduate Scientific Research Innovation Project of Xiangtan University (grant number XDCX2021B101).
Corresponding author: \texttt{jpzhang@xtu.edu.cn} (J. Zhang).}
\thanks{A. Liu is with the School of Mathematics and Computational Science, Xiangtan University, and Hunan Key Laboratory for Computation and Simulation in Science and Engineering, Xiangtan 411105, China (\texttt{202121511147@smail.xtu.edu.cn}).}
\thanks{X. Fan is with the School of Mathematics and Computational Science, Xiangtan University, and the Key Laboratory of Intelligent Computing \& Information Processing of the Ministry of Education, Xiangtan 411105, China (\texttt{fanxiaohong@smail.xtu.edu.cn}).}
\thanks{Y. Yang and J. Zhang are with the School of Mathematics and Computational Science, Xiangtan University, and the National Center for Applied Mathematics in Hunan, Xiangtan 411105, China. (\texttt{yangyinxtu,\;jpzhang@xtu.edu.cn}).}
}

\markboth{Journal of \LaTeX\ Class Files,~Vol.~xx, No.~xx, September~2023}%
{How to Use the IEEEtran \LaTeX \ Templates}
\maketitle
\begin{abstract}
The problem of phase retrieval (PR) involves recovering an unknown image from limited amplitude measurement data and is a challenge nonlinear inverse problem in computational imaging and image processing. However, many of the PR methods are based on black-box network models that lack interpretability and plug-and-play (PnP) frameworks that are computationally complex and require careful parameter tuning. To address this, we have developed PRISTA-Net, a deep unfolding network (DUN) based on the first-order iterative shrinkage thresholding algorithm (ISTA). This network utilizes a learnable nonlinear transformation to address the proximal-point mapping sub-problem associated with the sparse priors, and an attention mechanism to focus on phase information containing image edges, textures, and structures. Additionally, the fast Fourier transform (FFT) is used to learn global features to enhance local information, and the designed logarithmic-based loss function leads to significant improvements when the noise level is low. All parameters in the proposed PRISTA-Net framework, including the nonlinear transformation, threshold parameters, and step size, are learned end-to-end instead of being manually set. This method combines the interpretability of traditional methods with the fast inference ability of deep learning and is able to handle noise at each iteration during the unfolding stage, thus improving recovery quality. Experiments on Coded Diffraction Patterns (CDPs) measurements demonstrate that our approach outperforms the existing state-of-the-art methods in terms of qualitative and quantitative evaluations. Our source codes are available at \emph{https://github.com/liuaxou/PRISTA-Net}.
\end{abstract}


\section{Introduction}
Optical sensors in real imaging systems tend to capture only the amplitude of a signal, while phase information is lost. The objective of the phase retrieval (PR) problem is to identify the original signal $\bm{x}\in\mathbb{R}^N$ from the amplitude measurements $\bm{y}\in\mathbb{R}^M$. This can be expressed mathematically as:
\begin{equation}
\bm{y}=\left| \bm{Ax}\right|+\bm{\omega},
\label{PR-prob}
\end{equation}
where $\bm{A}\in\mathbb{C}^{M\times N}$ represents the forward measurement matrix and $\bm{\omega}\in\mathbb{R}^M$ denotes the measurement noise. The PR model has been extensively used in a variety of areas, including diffraction imaging \cite{1}, electron microscopy \cite{36}, optical microscopy \cite{43}, astronomy \cite{40}, X-ray crystallography \cite{44}, holography imaging \cite{41}, ptychography\cite{42}, fringe projection profilometry\cite{57} and super resolution\cite{58}.
 
The earliest algorithms to solve the PR problem (\ref{PR-prob}) were based on the alternating projection technique. For example, the Gerchberg-Saxton (GS) \cite{1} algorithm iteratively projects between the constraint sets in the image and frequency domains to recover the missing phase information from the amplitude measurements. Fienup \cite{2} extended the GS algorithm by introducing nonnegativity and support constraints, as well as replacing the amplitude constraints in the image domain. However, these methods are sensitive to noise. Subsequently, several improved algorithms based on Fienup's work were developed, such as hybrid PR \cite{3}, RAAR \cite{4}, and OSS \cite{50}. Researchers have employed convex optimization theory to propose algorithms such as PhaseLift \cite{5} and PhaseCut \cite{6} that transform non-convex PR problems into convex optimization problems using matrix lifting. Nevertheless, their high computational cost makes them difficult to apply in practical scenarios. Wirtinger flow (WF) \cite{26} initially obtains an accurate estimate using spectral methods and then updates it through a stochastic gradient descent based on Wirtinger derivatives. The truncated Wirtinger flow (TWF) \cite{27} improves on WF by adopting the Poisson loss function and preserving well-behaved measurements through a truncation threshold. This data-tuning process results in a more stringent initial guess and a more stable search direction. However, these WF-based methods mainly focus on the generalized PR problem, where the elements of the measurement matrix follow a Gaussian distribution. This significantly restricts their practical applications, and these algorithms have stringent convergence requirements.

Following the influence of compressive sensing, many PR algorithms have been developed that incorporate sparse prior information, such as dictionary learning \cite{7} and total variation \cite{8}. However, these regularization techniques rely on manually designed priors and require manual tuning of hyperparameters, and the imposed prior knowledge may not take full advantage of the available image data. Additionally, these methods often require a significant amount of computational costs.

Recently, deep neural networks (DNNs) have been successful in a variety of imaging tasks \cite{59,60,QZuo2021,HZhu2022}, and researchers have used end-to-end black-box DNNs for PR problems. DeepMMSE\cite{25} approximates the estimation of the minimum mean square error by using dropout-based model averaging. Işıl et al.\cite{51} combined two DNN modules with the HIO algorithm to iteratively improve the quality of the reconstruction. However, black-box networks are not interpretable.

In addition, some researchers have combined the Plug-and-Play (PnP) framework \cite{28} with deep denoisers. For example, prDeep \cite{9} combines the Regularization by Denoising (RED) framework \cite{10} with PR models and incorporates a pre-trained DnCNN denoiser into the PnP framework. Shi et al. \cite{11} have imposed sparse priors on unknown images using a tight frame \cite{45} and included a pre-trained deep denoiser \cite{46} in the model for high-resolution diffraction imaging. Chen et al. \cite{12} designed a prior denoising based on complex-valued neural networks in the Gabor domain \cite{47}, and inserted the pretrained denoiser into the RED framework. However, the success of these methods is contingent on the correlation between the image distribution used for pretraining and the distribution of the target image. Furthermore, the need for cumbersome parameter tuning further restricts their practical applicability.
 
Generative models have been demonstrated to be able to capture high-dimensional image distributions, leading to better performance than hand-crafted priors. Hyder et al. \cite{38} combined an alternating optimization strategy with a pretrained generative prior to solve the noiseless compressed PR problem. Shamshad et al. \cite{39} proposed a framework to regularize the PR problem through deep generative priors by using a gradient descent algorithm. However, the reconstruction results of such algorithms are usually restricted to the range of generative models, and inference is slow because of layer-wise optimization.
  
The Deep Unfolding Network (DUN) has been used to truncate and unfold traditional optimization algorithms into DNNs, combining the interpretability of traditional optimization algorithms with the fast inference capability of neural networks. It has been widely applied to various imaging inverse problems \cite{29,30,31,32,17,33,52,34,55}. Researchers have applied DUN to PR problems, such as TFPnP \cite{13}, which inserts pre-trained deep denoisers into PnP frameworks and uses deep reinforcement learning strategies to adjust parameters. Yang et al. \cite{14} proposed a dynamic proximal-point unfolding network that can adapt to different imaging conditions and process multiple compressed imaging models with a single trained model, allowing adjustment of dynamic parameters during the inference stage. Yang et al. \cite{35} replaced the image domain projection operator in HIO \cite{2} with a prior projection module, although the network structure is complex and the inference time is long. Zhang et al. \cite{36} used a neural network constructed with a complex value of UNet \cite{37} to replace the image domain and frequency domain update functions in the traditional GS algorithm, but this approach is limited to specific imaging applications. To reduce the cost of collecting real images in practical applications, some researchers have trained neural networks using unpaired data. For example, Cha et al. \cite{15} designed an unsupervised learning method based on traditional alternating projection and PhaseCut algorithms, which can be trained on unpaired data, but this method is limited to Fourier measurements.
  
There are still three main shortcomings in PR algorithms: (1) The existing black-box networks algorithms for PR lack interpretability. (2) The existing PnP-based algorithms which use pre-trained denoisers have problems such as performance degradation due to image domain shift and complex parameter tuning. (3) Due to the presence of modulus operators, PR is a typical non-convex problem. End-to-end trained PR algorithms based on DUNs usually depend on traditional alternating projection algorithms. Since traditional convex optimization algorithms can only find local optimization, there are few works that combine first-order iterative optimization algorithms with PR and map them into DUNs. Based on the above observations, this paper makes the following contributions:
\begin{enumerate}
	\item[(1)] We propose a novel PRISTA-Net that combines model-driven and data-driven approaches, as a powerful tool for PR tasks. This network has the benefits of both traditional algorithms, which are interpretable, and deep learning, which is highly effective in learning;
	\item[(2)] We utilize a well-designed DNN to substitute the nonlinear transformation operator in the proximal-point mapping subproblem of ISTA. Specifically, we use convolutional processing in both the spatial and frequency domains to capture local and global information. Furthermore, the attention mechanism is used to focus the network on the phase information that contains image edges, textures, and structures, thus aiding phase retrieval.
	\item[(3)] Compared to PnP-based algorithms, all parameters of the proposed PRISTA-Net can be learned end-to-end without complex parameter tuning. Moreover, our logarithmic-based loss function can further improve the results, particularly in low noise levels. Comprehensive experiments demonstrate that our PRISTA-Net outperforms existing state-of-the-art algorithms while preserving desirable computational complexity.
\end{enumerate}

\section{Methodology}
\begin{figure*}[htbp]
\centering
\includegraphics[width=1.0\linewidth]{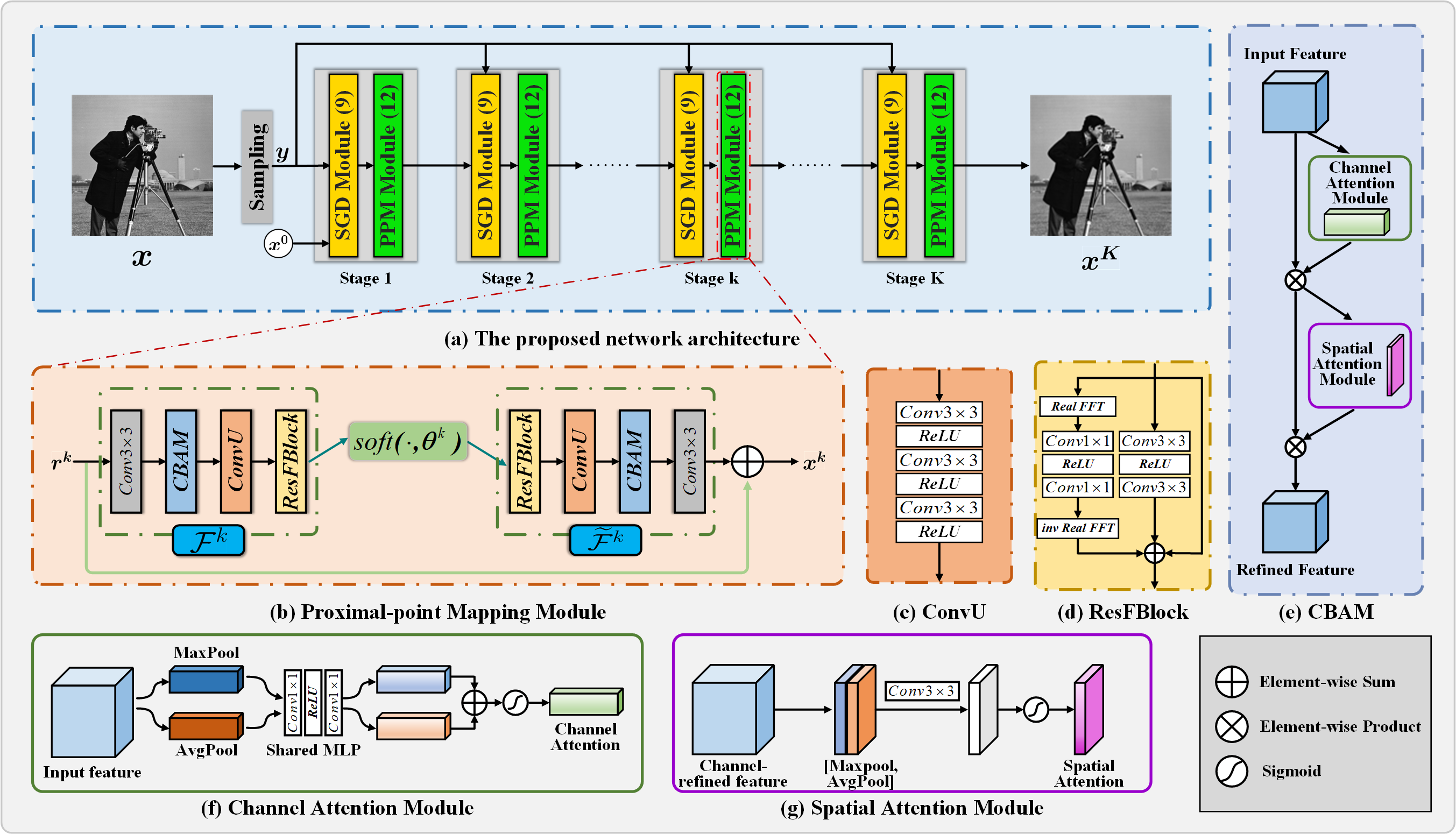}
\caption{The overall architecture of the proposed PRISTA-Net.}
\label{PR-fig1}
\end{figure*}

We start by reviewing the Coded Diffraction Patterns (CDPs) measurement model and the traditional ISTA algorithm to solve the PR regularization problems. ISTA algorithm has become a widely used composite optimization technique in mathematical methods for image processing in the last two decades. In addition, we provide a thorough explanation and present a comprehensive analysis of resolving the PR problem by training an unfolded ISTA framework. The architecture of the proposed PRISTA-Net is illustrated in Fig. \ref{PR-fig1}.

\subsection{Measurement and Noise Model}
In order to reconstruct interesting objects, different spatial light modulators (SLMs) are used to acquire the desired frequency information. The CDPs measurement model \cite{22} involves illuminating the object with a coherent source and then modulating its phase with random patterns using SLMs. According to Fraunhofer's diffraction equation, the optical field of the detector can be accurately modeled by the 2D Fourier transformation of the object of interest through the propagation of a wide field. It is well known that photographic plates, CCDs, and other light detectors can only measure the intensity of the light field and the interference of noise \cite{13}. Therefore, the measurement model is formulated as follows.
\begin{equation}\label{14}
\bm{y}^2=|\bm{Ax}|^2+\bm{\omega},
\end{equation}
where $\bm{A}=[(\bm{FD}_1)^T,(\bm{FD}_2)^T,\cdots,(\bm{FD}_J)^T]^T$, $\bm{F}$ is 2D Fourier transform matrix, and $\{\bm{D}_j\}_{j=1}^J$ are the illumination masks used to simulate SLMs. In previous work, non-compressive CDPs were typically generated using uniform masks, whereas binary masks were used to create compressive CDPs. In this study, we use uniform masks, where $\bm{D}_j$ is a diagonal matrix with elements randomly sampled from the unit circle of the complex plane. As suggested in \cite{25}, the number of masks is $J=1,2,3,4$. $\bm{\omega}$ is shot noise (following a Poisson distribution) with $\bm{\omega}\sim N(0,\alpha^2 \bm{Ax} ^2)$. $\alpha$ is used to adjust the signal-to-noise ratio (SNR), with a larger $\alpha$ indicating a lower SNR. More details can be found in \cite{22,9}.
\subsection{Iterative PR}
As well known, a generalized PR approach based on regularization priors can be expressed mathematically as follows:
\begin{equation}\label{1}
\arg\min_{\bm{x}\in\mathbb{R}^N}f(\bm{x})+\gamma g(\bm{x}),
\end{equation}
where $f(\bm{x})$ is the data fidelity term, $g(\bm{x})$ is the regularization term used to enforce data priors, and the regularization parameter $\gamma$ is used to balance $f(\bm{x})$ and $g(\bm{x})$. According to ISTA \cite{16}, the $\ell_1$ norm of $\bm{x}$ in a certain transform domain $\bm{\Psi}$ is used to impose the prior sparsity, that is, $g(\bm{x})=\Vert \bm{\Psi} \bm{x}\Vert_1$. 

Recently, image inverse problems have focused on a particular application that involves a Bayesian approach. The data fidelity term $f(\bm{x})$ is proportional to the negative log-likelihood function. Additionally, a link between least squares and post-prior probability has been established in \cite{56}.
The negative log-likelihood of $w\sim N(0,\sigma^2\bm{I})$ in (\ref{14}) is expressed as $-\log p(\bm{y}| \bm{x})\propto \Vert \bm{y}^2- |\bm{Ax}| ^2\Vert^2$. However, the Poisson log-likelihood function, which is of particular interest in this paper, has been found to be less effective empirically than amplitude loss $\frac{1}{2}\fan[\bm{y}- |\bm{Ax}| ]^2_2$ \cite{53,9,54}. Therefore, we substitute the amplitude loss for the data fidelity term $f(x)$ in (\ref{1}).
\begin{equation}\label{15}
\arg\min_{\bm{x}\in\mathbb{R}^N}\frac{1}{2}\fan[\bm{y}-|\bm{Ax}|]^2_2+\gamma \Vert \bm{\Psi} \bm{x}\Vert_1.
\end{equation}
 
Using formal calculations, e.g., the second-order Taylor expansion at $\bm{x}^{k-1}$ which is analogous to our explanation involving the first-order gradient of $f(\bm{x})$, we can derive the following least square approximation:
\begin{equation}\label{2}
f(\bm{x})\approx f^k(\bm{x})=\frac{1}{2\eta^k}\fan[\bm{x}-(\bm{x}^{k-1}-\eta^k\partial f(\bm{x}^{k-1}))]^2_2+c,
\end{equation}
where $c$ is a constant which is independent of $\bm{x}$. 
We recall that the solution of (\ref{15}) is equally obtained by the proximal-point gradient descent algorithm with two-step iterations, that is,
\begin{equation}
\bm{r}^k=\bm{x}^{k-1}-\eta^k\partial f(\bm{x}^{k-1})
\label{3}
\end{equation}
and
\begin{equation}
\begin{split}
\bm{x}^k&=\textbf{Prox}_{\lambda^k\Vert \bm{\Psi} (\cdot)\Vert_1}(\bm{r}^k)\\
&=\argm[\bm{x}]\frac{1}{2}\fan[\bm{x}-\bm{r}^k]^2_2+\lambda^k \Vert \bm{\Psi} \bm{x}\Vert_1,
\end{split}
\label{5}
\end{equation}
where $\eta^k$ is the step-length which is obtained by line search in traditional composite optimization algorithms, and $\lambda^k=\eta^k \gamma$.

Solving the proximal-point mapping problem (\ref{5}) accurately and efficiently is essential for traditional ISTA. If $\bm{\Psi}$ is a wavelet transform matrix that satisfies the orthogonality property \cite{18}, then $\textbf{Prox}_{\lambda\Vert \bm{\Psi} (\cdot)\Vert_1}(\bm{r})=\bm{\Psi}^T\textbf{soft}(\bm{\Psi} \bm{r},\lambda)$ \cite{17}. However, the conditions required for this transformation matrix are often too strict, leading to poor reconstruction performance. Furthermore, $\bm{\Psi}$ is usually chosen as a fixed transformation such as DCT \cite{48}, wavelet transform, or gradient operators \cite{19}. These approaches have a strong theoretical basis, but usually require manual parameterization and a very high computational cost. For example, the selection of the step length and stopping iteration in (\ref{3}) is essential. Furthermore, determining the nonconvex feasible set $\mathcal{X}$ of the optimization problem (\ref{15}), that is, if $x\in \mathcal{X}$, meaning $x$ satisfies $y=|Ax| +\omega$, then $-x\in\mathcal{X}$, but $x+(-x)\notin\mathcal{ X}$, it is very likely that the classical ISTA algorithm will converge to a local optimal or suboptimal solution. Therefore, we do not provide a rigorous convergence proof, but it has been demonstrated experimentally that the algorithm with this iterative idea can achieve better results, that is, it can be assumed to converge to a good solution. 

\subsection{The proposed PRISTA-Net architecture}
The concept of PRISTA-Net is to combine the interpretability of first-order proximal gradient optimization algorithms with the strong expressive power of neural networks. This is done by truncating the ISTA approximations and mapping them into a deep network structure with a fixed number of stages. Similarly to ISTA-Net \cite{17}, each stage of PRISTA-Net consists of two modules: the Subgradient Descent Module (SGD module) and the Proximal-point Mapping Module (PPM module).
 
\textbf{Subgradient Descent (SGD) Module:} 
The CDPs measurement model (\ref{14}) can present a challenge, as the solution $\bm{x}$ may not be unique. To address this, a common approach is to solve the nonlinear least-squares problem using first-order gradient methods, such as the steepest gradient descent or trust region method. In this paper, however, we will use the subgradient descent method.

The purpose of this step is to measure the direction of change at the current point (Fig. \ref{PR-fig1} (a)) using equation (\ref{3}), which is essential for updating the iteration point. The gradient indicates the direction of the function that increases the most in the vicinity of the current point. To reduce the value of the objective function, we move in the opposite direction of the gradient.

It should be noted that $\partial f(\bm{x}^{k-1})$ is the gradient of the data fidelity term $f(\bm{x})$ at $\bm{x}^{k-1}$. Since $f(\bm{x})=\frac{1}{2}\fan[\bm{y}- \bm{|Ax|} ]^2_2$ is not differentiable, we use the sub-gradient of $f(x)$ instead of the gradient. This is expressed as 
\begin{equation}\label{7} 
\bm{A}^H(\bm{Ax}-\bm{y}\circ \frac{\bm{Ax}}{ |\bm{Ax}| })\in\partial_{\bm{x}}\frac{1}{2}\fan[\bm{y}- \bm{|Ax|}]^2_2, 
\end{equation} 
where the symbol $\circ$ denotes the Hadamard product that is element-wise multiplication. $\bm{A}^H$ is the conjugate transpose of $\bm{A}$. $\partial_{\bm{x}}\frac{1}{2}\fan[\bm{y}- \bm{|Ax|} ]^2_2$ is the subdifferential set of $f(\bm{x})$ at $\bm{x}$. Therefore, the SGD module can be formulated as 
\begin{equation}\label{8} 
\begin{aligned} \bm{r}^k=&\mathcal{SGD}(\bm{x}^{k-1},\eta^k,\bm{y},\bm{A})\\ 
=&\bm{x}^{k-1}-\eta^k\bm{A}^H\left(\bm{A}(\bm{x}^{k-1})-\bm{y}\circ \frac{\bm{A}(\bm{x}^{k-1})}{ |\bm{A}(\bm{x}^{k-1})|}\right), 
\end{aligned} 
\end{equation} 
where $\eta^k$ is the learnable step size.
 
\textbf{Proximal-point Mapping (PPM) Module:} Inspired by the traditional ISTA proximal-point mapping module whose role is to threshold $\bm{r}^k$ to obtain a sparse solution, the PPM module aims
to compute $\bm{x}^k$ from $\bm{r}^k$ by comparing each element of $\bm{r}^k$ to a threshold value and reducing those that exceed it, thus setting insignificant elements to zero. This allows for signal sparsity and resolution of the sparsity problem. Generally, the threshold is determined by cross-validation or experience. The selection of the threshold value has an essential effect on the quality of the final sparse solution and the speed of convergence, while our method can adapt the threshold value through end-to-end learning.

Similarly to \cite{17}, we employ a parameterized learnable nonlinear transformation $\mathcal{F}$ instead of $\bm{\Psi}$ to impose sparse priors, as illustrated in Fig. \ref{PR-fig1} (b). According to Theorem 1 from \cite{17}, (\ref{5}) can be expressed as:
\begin{equation}\label{10}
\bm{x}^k=\arg\min\limits_{\bm{x}}\frac{1}{2}\Vert \mathcal{F}(\bm{x})-\mathcal{F}(\bm{r}^k)\Vert^2_2+\theta^k\Vert \mathcal{F}(\bm{x})\Vert_1,
\end{equation}
where $\theta^k$ is the learnable threshold value related to $\lambda^k$. Therefore, a closed-form solution for $\mathcal{F}(\bm{x}^k)$ can be derived as follows:
\begin{equation}\label{11}
\mathcal{F}(\bm{x}^k)=\textbf{soft}(\mathcal{F}(\bm{r}^k),\theta^k).
\end{equation}
Inspired by the invertibility property of the transform operator in traditional iterative algorithms, and denoting the left inverse $\widetilde{\mathcal{F}}$ of $\mathcal{F}$, we have
\begin{equation}\label{12}
\bm{x}^k=\widetilde{\mathcal{F}}(\textbf{soft}(\mathcal{F}(\bm{r}^k),\theta^k)).
\end{equation}

As illustrated in Fig. \ref{PR-fig1} (b), our nonlinear transformation function $\mathcal{F}$ is composed of four parts. Initially, the intermediate result $\bm{r}^k$ is subjected to a $3\times3$ convolution to ensure high-throughput information transfer and to avoid information loss. Subsequently, the Convolutional Block Attention Module (CBAM) \cite{20} is used to improve the model's ability to focus on relevant information. The channel attention module and the spatial attention module are used to adaptively learn the weights of different channels and locations, so that the model can concentrate on more important channel and spatial information. By combining channel and spatial attention, the CBAM module can extract features with stronger characterization ability and richer contextual information, allowing the network to focus on phase information such as edges, textures, and structures of images. The features are then further refined by passing through convolutional units (ConvU), which consists of three $3\times3$ convolutions with a Relu function, to capture more fine-grained details. Finally, the features are processed by the Fourier Residual Block (ResFBlock) \cite{21}, which enhances the representation ability for both low-frequency and high-frequency by using global information and local details. 

It is worth noting that ResFBlock has an extra frequency branch in comparison to the traditional Residual Block, which transforms features into the frequency domain. The FFT, which captures global receptive fields (each element in the frequency domain is associated with all elements in the image), provides complementary global information learning to the local details learned by the classical Residual Block. Subsequently, the improved features are processed by the soft-thresholding operator, and the insignificant information is filtered to achieve signal sparsification. The left inverse $\widetilde{\mathcal{F}}$ is structured to be symmetric with $\mathcal{F}$. It consists of ResFBlock, ConvU, CBAM and a 3×3 convolution which recovers the corrected high-throughput information to a promising image.

\begin{figure*}[htbp]
\centering
\includegraphics[width=1.0\linewidth]{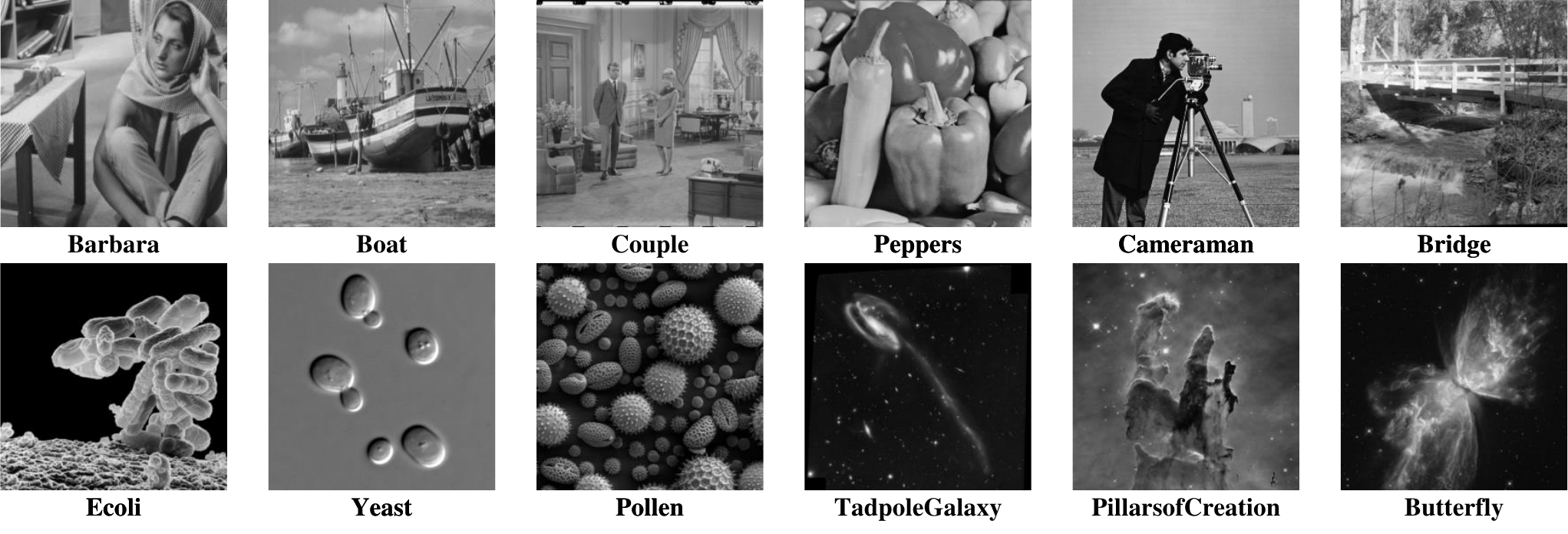}
\caption{Commonly used test dataset in PR. Top row: NT-6 and bottom row: UNT-6. They follow different distributions.}
\label{fig:gt}
\end{figure*}
 
To increase the network's capacity for feature representation, facilitate optimization of the network, and prevent vanishing or exploding gradients during training, we designed the entire PPM module as a residual structure.
\begin{equation}
\begin{aligned} 
\bm{x}^k=&\mathcal{PPM}_{\theta^k}(\bm{r}^k,\mathcal{F}^k,\widetilde{\mathcal{F}}^k)\\
=&\bm{r}^k+\widetilde{\mathcal{F}}^k(\textbf{soft}(\mathcal{F}^k(\bm{r}^k),\theta^k)).
\end{aligned}
\end{equation}

\subsection{Loss Function}
Given a training dataset $\{(y_i,x_i)\}_{i=1}^{N_s}$, PRISTA-Net takes the measurements $y_i$ and the initialization $x^0$ as input and produces the final reconstruction result $x^K_i$ in the $K$-th stage. $\mathcal{N}_s$ is the number of samples in the training dataset and $K$ is the number of stages. The loss function is used to guide the learning of network parameters by minimizing the distance between the reconstruction prediction and the ground truth, and the most commonly used are $l_1$ and $l_2$. The total loss is the logarithm of the sum of the mean square error between the reconstruction result and the ground truth at each stage, i.e. 
\begin{equation}\label{13}
\mathcal{L} = \log(\sum_{k=1}^{K}\mathcal{L}^k),\;\mathcal{L}^k=\frac{1}{\mathcal{N}_s\mathcal{N}}\sum_{i=1}^{\mathcal{N}_s}\fan[x_i^k-x_i]^2_2,
\end{equation}
where $\mathcal{N}$ is the size of each sample. The reason for using this loss is that when the mean square error drops to less than $1$, the dynamic range of the error can be further amplified by using the logarithmic function, which allows the network to continue in the descending direction and avoid the current local minimum. For example, when $K=1$, let $\Theta$ denote all the learnable parameters, $\mathcal{L}_\Theta=\log(\mathcal{L}^1_\Theta)$, then $\frac{\partial\mathcal{L}_\Theta}{\Theta}=\frac{\partial\mathcal{L}^1_\Theta}{\partial\Theta}$$\frac{1}{\mathcal{L}^1_\Theta}$. Therefore, when $\mathcal{L}^1_\Theta\le 1$, the dynamic range of $\frac{\partial\mathcal{L}_\Theta}{\Theta}$ increases.

\subsection{Parameters and initialization}
The learnable parameters of the proposed architecture are represented as $\bm{\Theta}=\{\eta^k,\theta^k,\mathcal{F}^k,\widetilde{\mathcal{F}}^k\}_{k=1}^K$, consisting of the step size $\eta^k$ in the SGD module, the threshold $\theta^k$ in the PPM module, and the non-linear transformation functions $\mathcal{F}^k$ and the inverse transformation $\widetilde{\mathcal{F}}^k$. All these parameters can be learned by minimizing the loss function (\ref{13}). The Xavier\cite{49} method is used to initialize the proposed network. $\eta^1$ and $\theta^1$ are initialized with 0.5 and 0.01, respectively. $K$ is set to $7$ and the number of convolution channels is set to 32 by default. Parameters among stages are not shared. Following PrDeep \cite{12}, the initial iteration input $x^0=1$.

\begin{figure*}[!htbp]
\subfigure[]{
\includegraphics[width=5.8cm,height=5cm]{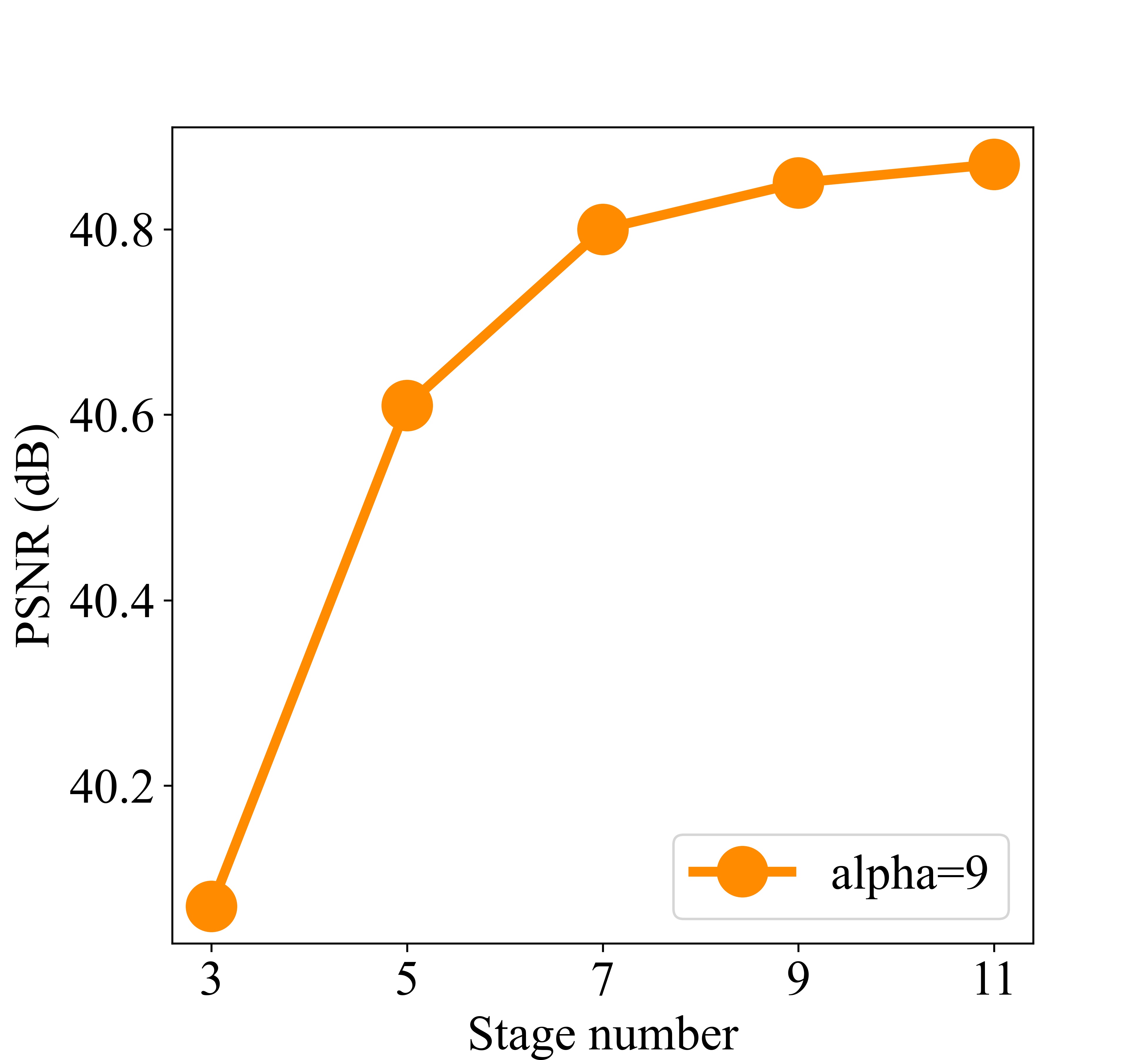} }
\subfigure[]{
\includegraphics[width=5.8cm,height=5cm]{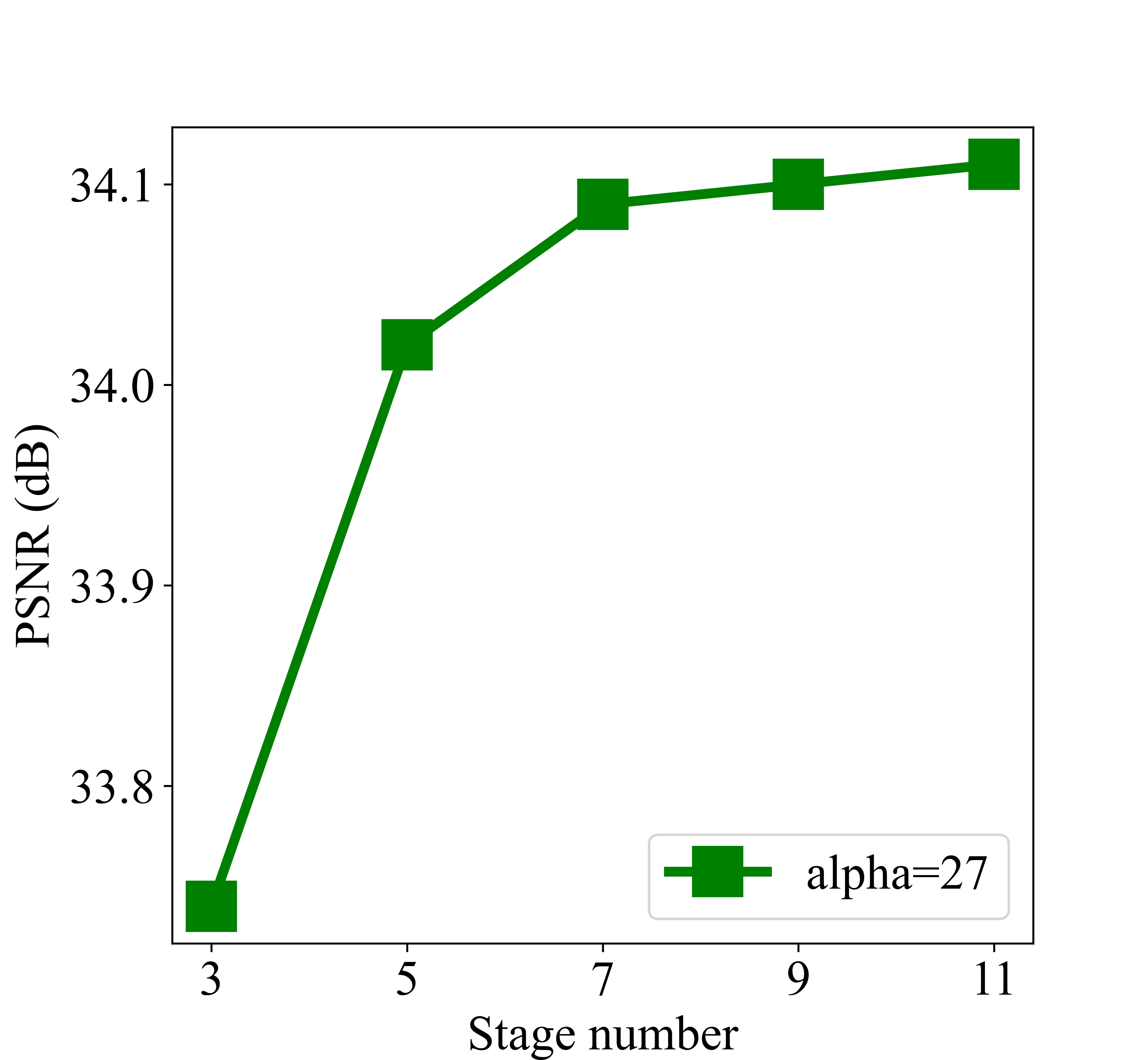} }	
\subfigure[]{
\includegraphics[width=5.8cm,height=5cm]{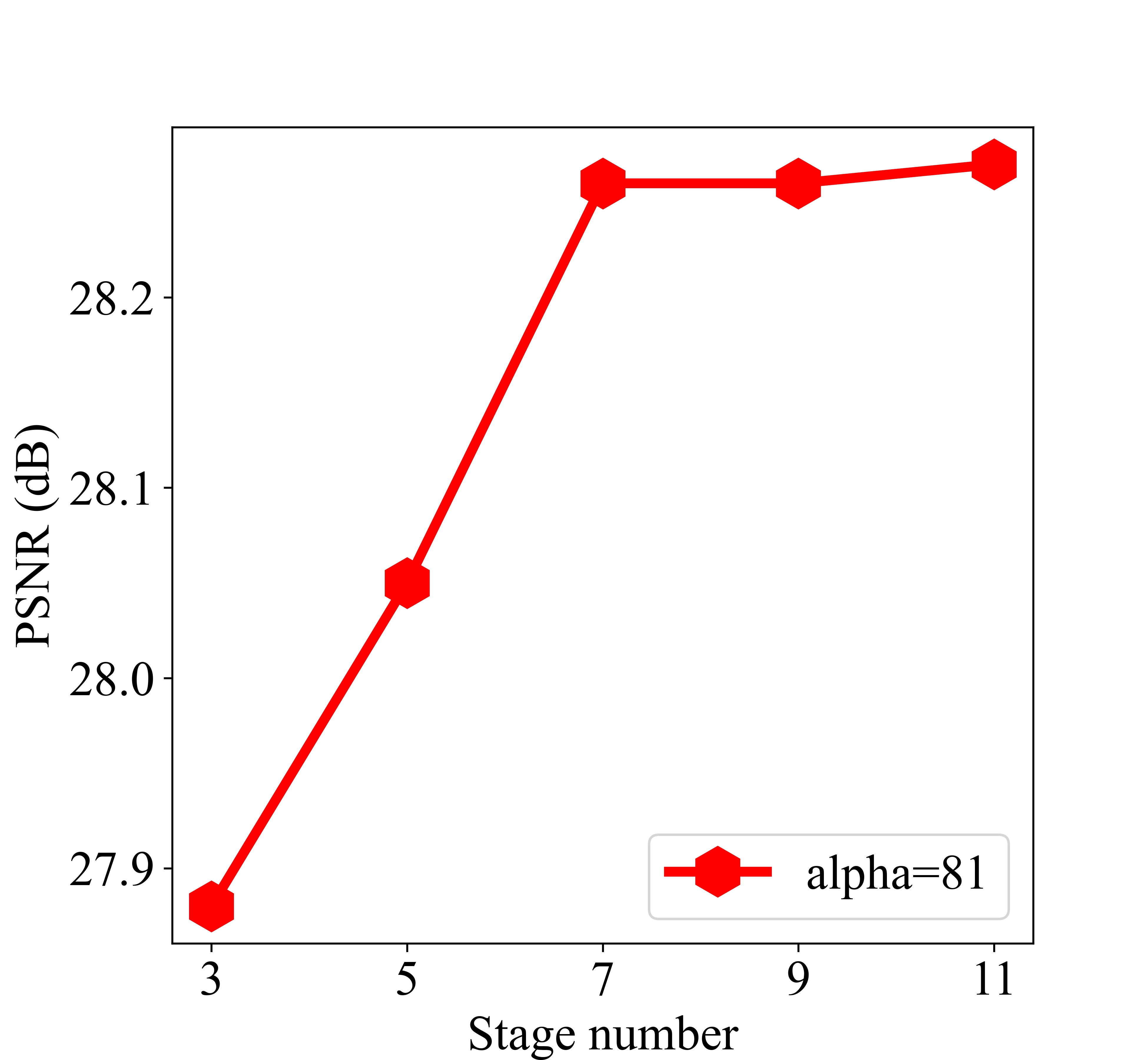} }	
\caption{Average PSNR (dB) results on the 128 × 128 test dataset with various stages using 4 uniform masks at different noise levels.}
\label{Phase_num}
\end{figure*}

\section{Experiments}
\subsection{Experimental Setup}
Our training set consists of 6,000 images from BSD400 \cite{23} and PASCAL VOC \cite{24}. Each image is resized to $128\times128$ to generate CDPs (\ref{14}). We follow PrDeep \cite{9} and use 6 natural and 6 unnatural images as test datasets, denoted NT-6 and UNT-6 (Fig.\ref{fig:gt}), with two scales: $128\times128$ and $256\times256$. The image pixel range is $[0,1]$, and similar to \cite{13}, we add possion noise by dividing the noise level by $255$ to fit the image pixel range. The noise level $\alpha$ is uniformly sampled from $\{9,27,81\}$ during training. To assess the practicality of the algorithm, we use different masks during the testing phase than during the training phase, but several compared algorithms are tested using the same masks, as in \cite{9}.

We employ Pytorch to implement our method and use the Adam optimizer with an initial learning rate of $0.001$. The learning rate is multiplied by a factor of $0.95$ every two epochs, and a total of $200$ epochs are trained with a batch size of $10$. The training is conducted on a workstation with an Intel(R) Xeon(R) Silver 4214 CPU and a Tesla V100-PCIE-32GB, taking approximately 7 hours to train PRISTA-Net with $K=7$ stages. The reconstruction results are evaluated using the average peak signal-to-noise ratio (PSNR) and the average structural similarity index measure (SSIM).

\begin{table}
\caption{Average psnr (dB) of reconstructions on the $128\times 128$ test dataset with different channel numbers using 4 uniform masks at different noise levels.}
\label{table7}
\setlength\tabcolsep{2pt}
\centering
\begin{tabular*}{\hsize}{@{}@{\extracolsep{\fill}}cllllll@{}}
\toprule[1.5pt]
\multirow{2}{*}{Channel} &\multicolumn{2}{c}{$\alpha=9$} &\multicolumn{2}{c}{$\alpha=27$} & \multicolumn{2}{c}{$\alpha=81$} \\ 
\cmidrule(lr){2-3}\cmidrule(lr){4-5}\cmidrule(lr){6-7}
&UNT-6 &NT-6 & UNT-6 & NT-6& UNT-6 & NT-6  \\
\midrule[0.7pt]
4&40.11& 38.68&33.84& 32.28&28.54& 26.27\\
8&40.30& 39.29&34.22& 32.59&28.85& 26.45\\
16&41.01& 39.89&34.74& 33.09&29.20& 26.93\\
32&\textbf{41.45}& \textbf{40.12}&\underline{34.90}& \underline{33.26}&\textbf{29.47}&\textbf{27.14}\\
64&\underline{41.38}&\underline{40.09}&\textbf{34.93}&\textbf{33.30}&\underline{29.45}&\underline{27.11}\\
\bottomrule[1.5pt]
\end{tabular*}
\end{table}

\begin{table}
\caption{Average psnr (dB) of reconstructions on the $128\times128$ test dataset with different loss functions using 4 uniform masks at different noise levels.}
\label{table6}
\setlength\tabcolsep{2pt}
\centering
\begin{tabular*}{\hsize}{@{}@{\extracolsep{\fill}}ccccccc@{}}
\toprule[1.5pt]
\multirow{2}{*}{Setting} &\multicolumn{2}{c}{$\alpha=9$}&\multicolumn{2}{c}{$\alpha=27$}&\multicolumn{2}{c}{$\alpha=81$} \\ 
\cmidrule(l){2-3}\cmidrule(l){4-5}\cmidrule(l){6-7}
&\makecell{UNT-6} & \makecell{NT-6} & \makecell{UNT-6} & \makecell{NT-6} & \makecell{UNT-6} & \makecell{NT-6}  \\
\midrule[0.7pt]
$w/\;\log$ (default)&\textbf{41.45}&\textbf{40.12}&\textbf{34.90}&\textbf{33.26}&\textbf{29.47}&\textbf{27.14}\\
$w/o\;\log$ &41.17&39.89&34.77&33.20&29.45&27.14\\
$\mathcal{L}^K$  &40.81&39.53&34.53&32.79&29.16&26.84\\
\bottomrule[1.5pt]
 \end{tabular*}
\end{table}

\begin{table*}
\caption{Average psnr (dB) of reconstructions on the 128$\times$128 test dataset using 4 uniform masks with different combinations of CBAM and ResFBlock at different noise levels.}
\label{table5}
\setlength\tabcolsep{2pt}
\centering
\begin{tabular*}{\hsize}{@{}@{\extracolsep{\fill}}ccccllllll@{}}
\toprule[1.5pt]
\multirow{2}{*}{Variant} &\multirow{2}{*}{CBAM} &\multirow{2}{*}{ResFBlock}&\multirow{2}{*}{Parameters}&\multicolumn{2}{c}{$\alpha=9$}&\multicolumn{2}{c}{$\alpha=27$}&\multicolumn{2}{c}{$\alpha=81$} \\ 
\cmidrule(lr){5-6}\cmidrule(lr){7-8}\cmidrule(lr){9-10}
& & & &UNT-6 & NT-6 & UNT-6 & NT-6 & UNT-6 & NT-6  \\
\midrule[0.7pt]
(a)&-&-&392693& 40.26 &39.27&34.19 &32.64 & 28.74&26.66\\
(b)&+&-&394737&41.19&39.89&34.69& 33.08&29.11& 26.87 \\
(c)&-&+&765429&41.36& 40.01&34.81& 33.21&29.35& 27.03\\
(d)&+&+&767473&\textbf{41.45}&\textbf{40.12}&\textbf{34.90}&\textbf{33.26}&\textbf{29.47}&\textbf{27.14}\\
\bottomrule[1.5pt]
\end{tabular*}
\end{table*}

\begin{table*}
\caption{Average psnr (dB) of reconstructions on the $128\times 128$ test dataset with different shared settings using 4 uniform masks at different noise levels.}
\label{table8}
\setlength\tabcolsep{2pt}
\centering
\begin{tabular*}{\hsize}{@{}@{\extracolsep{\fill}}cccllllll@{}}
\toprule[1.5pt]
\multirow{2}{*}{Variant}&\multicolumn{2}{c}{Shared setting} &\multicolumn{2}{c}{$\alpha=9$}&\multicolumn{2}{c}{$\alpha=27$}&\multicolumn{2}{c}{$\alpha=81$} \\ 
\cmidrule(lr){2-3}\cmidrule(lr){4-5}\cmidrule(lr){6-7}\cmidrule(lr){8-9}
& ResFBlock&CBAM &\makecell{UNT-6} & \makecell{NT-6} & \makecell{UNT-6} & \makecell{NT-6} & \makecell{UNT-6} & \makecell{NT-6}  \\
\midrule[0.7pt]
(a)& $\checkmark$&$\checkmark$ &41.33&39.97&34.70&33.18&29.25&27.03\\
(b)&$\checkmark$&$\times$&41.24&39.95&34.74&31.10&29.21&26.96\\
(c)&$\times$&$\checkmark$&41.39&40.01&34.81&33.18&29.30&26.99\\
(d)&$\times$&$\times$&\textbf{41.45}&\textbf{40.12}&\textbf{34.90}&\textbf{33.26}&\textbf{29.47}&\textbf{27.14}\\
\bottomrule[1.5pt]
\end{tabular*}
\end{table*}

\subsection{Intra-method evaluation}
We initially perform experiments to evaluate the effect of different components of our PRISTA-Net on reconstruction quality, such as the number of stages $K$, the number of channels, the loss function used, an analysis of the CBAM and ResFBlock modules, as well as various shared settings of PRISTA-Net.

\subsubsection{The Impact of Stage Number}
Firstly, we investigate the relationship between the number of stages and the average PSNR. Fig.\ref{Phase_num} shows the average PSNR curves in the $128\times128$ test dataset with respect to various stage numbers using 4 masks at different noise levels. 

From the results in Fig. \ref{Phase_num}, we observe that the average PSNR increases gradually with the number of stages $K$ and stabilizes after $K\geq7$. Therefore, taking into account the balance between network complexity and reconstruction performance, we set the default number of stages as $K=7$.

\subsubsection{The Impact of Channel Numbers}
We investigate the effect of the number of channels in the network by varying it from 4 to 64 and comparing the average PSNR at different noise levels using $4$ masks on the $128\times128$ test dataset. Table (\ref{table7}) shows that when the number of channels is increased from 4 to 32, there is a significant improvement in PSNR. However, when the number of channels is set to 64 and $\alpha=9,81$, the performance decreases slightly. 

To avoid overfitting, we have limited the number of channels in our network to 32, which is a good balance between complexity and reconstruction performance.

\subsubsection{The Effectiveness of Loss Function}
In this part, we also assess the effectiveness of our designed loss function by retraining the model with different loss functions and comparing their performance. We measure the average PSNR on the $128\times128$ test dataset using $4$ masks at different noise levels, and present the results in Table (\ref{table6}), where "$w/ \log$" denotes for using the designed logarithmic loss, "$w/o \log$" for without the designed logarithmic loss, and "$\mathcal{L}^K$" for using the $\mathcal{L}^K$ function as the loss.
 
Observing the results, we can see that the designed logarithmic loss yields considerable enhancements for various distributions of test images when the noise level is low. However, when the noise level is high, the improvements are restricted. Moreover, when only the mean square error between the output of the last stage and the ground truth is used as a loss function, the quality of the reconstruction decreases drastically. Thus, using the designed logarithmic loss has a certain effect on improving PR results.

\subsubsection{The Effectiveness of ResFBlock and CBAM}
To explore the effectiveness of the ResFBlock and CBAM modules, we evalute the average PSNR using $4$ masks in the $128\times128$ test dataset and analyze the corresponding increase in the number of parameters caused by these modules. The results are shown in Table \ref{table5}.

The baseline variant (a) does not use the ResFBlock and CBAM modules at each stage, and Table \ref{table5} shows that variant (b) with CBAM yields an average improvement of 0.52 dB compared to variant (a) with different datasets and noise levels. This indicates that CBAM's attention mechanism allows PRISTA-Net to extract more phase information that contains image details. The variant (c) uses the ResFBlock module in each stage. Although the ResFBlock module has fewer $3\times3$ convolutional layers than the ConvU module, the concatenation of real and imaginary parts in the frequency domain branch leads to a larger number of parameters, making the total number of parameters almost double that of variant (a). However, due to the complementary fusion of local and global information, variant (c) achieves a higher reconstruction quality and an average improvement of 0.67 dB compared to variant (a). This improvement is particularly noticeable in scenarios with lower noise levels. The variant (d) is the default version of the proposed PRISTA-Net. Compared to variant (c), it further incorporates the CBAM module at each stage, taking advantage of its attention mechanism to enhance the features. As a result, the reconstruction quality is further improved and yields an average 0.76 dB improvement compared to variant (a). These ablation experiments demonstrate the effectiveness of the ResFBlock and CBAM modules in PRISTA-Net.

\subsubsection{The  Impact of Module Sharing Configurations}
Here, we also show the flexibility of our proposed framework by performing several variants of PRISTA-Net with different shared settings between stages. Comparing the average PSNR in the $128\times128$ test dataset at different noise levels, the results are presented in Table \ref{table8}. We observe that PRISTA-Net achieves the best performance when using the default unshared version (d). Performance decreases when ResFBlock or CBAM shares parameters at all stages with less flexible. To achieve better results, we decided not to share the parameters among the more adaptable stages. 

Through the above ablation studies, we have analyzed the impact of convolution processing in both the spatial and frequency domains, the attention mechanism, and the designed logarithmic-based loss function on model performance by using various comparative experiments. We attribute the superiority of the default version of the proposed PRISTA-Net to the complementary learning of local and global information, the attention to the high-dimensional features containing image edge, texture, and structure, and the considerable improvement of the designed logarithmic-based loss when the noise level is low.

\begin{table*}
\caption{Average psnr (dB) /average ssim of PR  reconstructions on the 128$\times$128  and $256\times 256$ test datasets using different numbers of uniform masks at different noise levels.}
\label{table1}
\setlength\tabcolsep{1pt}
\centering
\begin{tabular*}{\hsize}{@{}@{\extracolsep{\fill}}cclllllll@{}}
\toprule[1.5pt]
\multirow{2}{*}{\makecell{\textbf{Size}}}&\multirow{2}{*}{\textbf{Sampling No.}} & \multirow{2}{*}{\textbf{Method}} & \multicolumn{2}{c}{$\bm{\alpha=9}$}&\multicolumn{2}{c}{$\bm{\alpha=27}$}&\multicolumn{2}{c}{$\bm{\alpha=81}$} \\ 
\cmidrule(r){4-5}\cmidrule(r){6-7}\cmidrule(r){8-9}
& & & \multicolumn{1}{c}{\textbf{UNT-6}}&\multicolumn{1}{c}{\textbf{NT-6}} & \multicolumn{1}{c}{\textbf{UNT-6}} &\multicolumn{1}{c}{\textbf{NT-6}} & \multicolumn{1}{c}{\textbf{UNT-6}}&\multicolumn{1}{c}{\textbf{NT-6}}  \\
\midrule[0.8pt]
\multirow{24}{*}{\rotatebox{90}{$\bm{128\times128}$}}&\multirow{6}{*}{$\bm{J=1}$}
&HIO \cite{2}&23.85/0.508&19.12/0.548&17.78/0.290&13.34/0.322&13.35/0.118&9.65/0.130\\
&&PrDeep \cite{9}&\underline{36.13}/0.948 &34.60/0.955&26.61/0.696&25.74/0.762&24.67/0.684&20.55/0.550 \\
&& TFPNP \cite{13}&35.99/\underline{0.950}&\textbf{34.77}/\textbf{0.958}&28.80/0.838&27.84/0.834&21.65/0.476&22.26/0.612 \\
&&DeepMMSE \cite{25}& 35.31/0.945&33.47/0.938&29.56/0.852&27.67/0.828&24.41/0.689&22.22/0.617 \\
&&PrComplex \cite{12}&35.70/0.944 &33.72/0.946&\underline{30.06}/\underline{0.854}&\underline{28.28}/\underline{0.859} & \underline{25.08}/\underline{0.727}&\underline{23.03}/\underline{0.681} \\
&&PRISTA-Net (Ours)&\textbf{36.38}/\textbf{0.954}&\underline{34.72}/\underline{0.956}&\textbf{30.90}/\textbf{0.877}&\textbf{29.16}/\textbf{0.876}&\textbf{26.00}/\textbf{0.744}&\textbf{23.72}/\textbf{0.689}\\
\\[-5pt]
&\multirow{6}{*}{$\bm{J=2}$}
&HIO \cite{2}&32.01/0.832&30.18/0.913&23.41/0.505&22.02/0.658&15.46/0.203&12.02/0.250\\
&&PrDeep \cite{9}&38.24/0.960 &\underline{36.87}/\underline{0.971}&31.86/0.895&30.47/0.907 &\underline{26.47}/0.775&24.00/0.737	 \\
&& TFPNP \cite{13}&36.06/0.933&34.70/0.959 &28.99/0.836&27.08/0.792&25.04/0.634&23.34/0.627\\
&&DeepMMSE \cite{25}& 38.00/\underline{0.966}&36.46/0.967 &31.04/\underline{0.903}&30.01/0.888 &25.72/0.762&23.70/0.698 \\
&&PrComplex \cite{12}&\underline{38.29}/0.965 &36.61/0.970 &\underline{32.21}/0.899&\underline{30.73}/\underline{0.908}  & 26.33/\underline{0.777}&\underline{24.23}/\underline{0.741}  \\
&&PRISTA-Net (Ours)&\textbf{39.03}/\textbf{0.972}&\textbf{37.47}/\textbf{0.975} &\textbf{32.81}/\textbf{0.916}&\textbf{31.18}/\textbf{0.915} &\textbf{27.64}/\textbf{0.797}&\textbf{25.31}/\textbf{0.757} \\
                \\[-5pt]
				&\multirow{6}{*}{$\bm{J=3}$}
                &HIO \cite{2}&33.90/0.892&31.20/0.948 &25.41/0.587&25.09/0.745 &17.24/0.273&14.37/0.358 \\
				&&PrDeep \cite{9}&39.31/0.963 &38.26/0.977 &\underline{33.24}/\underline{0.921}&\underline{31.99}/\underline{0.930} &\underline{27.19}/0.801&\underline{25.28}/\underline{0.788} \\
				&& TFPNP \cite{13}&\underline{40.02}/0.977&\underline{38.49}/\underline{0.980} &30.20/0.867&28.23/0.830  &25.28/0.688&23.47/0.663  \\
				&&DeepMMSE \cite{25}& 39.11/\underline{0.978}&38.20/0.976 &32.29/0.921&31.31/0.912 &26.52/0.791&24.64/0.749  \\
				&&PrComplex \cite{12}&39.55/0.972 &38.18/0.977  &33.20/0.914&31.96/0.926  &27.15/\underline{0.808}&25.24/0.786  \\
				&&PRISTA-Net (Ours)&\textbf{40.37}/\textbf{0.978}&\textbf{38.98}/\textbf{0.982} &\textbf{33.97}/\textbf{0.932}&\textbf{32.35}/\textbf{0.932} &\textbf{28.61}/\textbf{0.835}&\textbf{26.35}/\textbf{0.796} \\
                \\[-5pt]
				&\multirow{6}{*}{$\bm{J=4}$}
                &HIO \cite{2}&35.07/0.918&31.88/0.957 &26.79/0.645&26.52/0.790 &18.31/0.320&16.02/0.433 \\
				&&PrDeep \cite{9}&40.06/0.966 &39.23/0.981 &\underline{34.11}/0.932&\underline{32.92}/\underline{0.942} &\underline{27.83}/0.821&\underline{25.97}/\underline{0.817} 	 \\
				&& TFPNP \cite{13}&40.32/0.974&38.35/0.975 &30.56/0.866&28.82/0.838 &26.84/0.783&25.66/0.769  \\
				&&DeepMMSE \cite{25}& 40.27/\underline{0.982}&\underline{39.29}/0.981 &33.03/\underline{0.935}&32.20/0.926 &26.69/0.809&25.18/0.771  \\
				&&PrComplex \cite{12}&\underline{40.54}/0.976& 39.25/\underline{0.982} &33.90/0.925&32.77/0.937   & 27.78/\underline{0.825}&25.86/0.807  \\
				&&PRISTA-Net (Ours)&\textbf{41.45}/\textbf{0.982}&\textbf{40.12}/\textbf{0.985} &\textbf{34.90}/\textbf{0.941}&\textbf{33.26}/\textbf{0.943} &\textbf{29.47}/\textbf{0.853}&\textbf{27.14}/\textbf{0.825} \\
    \midrule[0.8pt]             
                \multirow{24}{*}{\rotatebox{90}{$\bm{256\times256}$}}&
				\multirow{6}{*}{$\bm{J=1}$}
                &HIO \cite{2}&23.74/0.443&18.75/0.440 &17.56/0.230&13.12/0.238 &13.57/0.093&9.38/0.098 \\
				&&PrDeep \cite{9}&37.43/0.942&\underline{35.89}/\underline{0.946}  &28.06/0.705&26.68/0.707 &25.04/0.629&21.14/0.490 \\
				&&TFPNP \cite{13}&37.10/0.939&35.26/0.942 &31.30/0.863&29.39/0.824 &25.84/0.683&\underline{25.69}/\underline{0.721} \\
                &&DeepMMSE \cite{25}&36.27/0.933&32.81/0.892&30.72/0.838&27.34/0.760&23.52/0.669&20.71/0.548  \\
				&&PrComplex \cite{12}&\underline{37.86}/\underline{0.950}&35.45/0.944 &\underline{31.93}/\underline{0.873}&\underline{30.15}/\underline{0.861} &\underline{26.99}/\underline{0.771}&24.76/0.712  \\
				&&PRISTA-Net (Ours)&\textbf{38.20}/\textbf{0.956}&\textbf{36.24}/\textbf{0.952} &\textbf{32.93}/\textbf{0.893}&\textbf{30.96}/\textbf{0.876} &\textbf{28.12}/\textbf{0.789}&\textbf{25.79}/\textbf{0.725} \\
				\\[-5pt] 
				&\multirow{6}{*}{$\bm{J=2}$}
                &HIO \cite{2}&32.17/0.798&29.71/0.870 &23.06/0.435&21.68/0.550 &15.34/0.160&11.65/0.183 \\
				&&PrDeep \cite{9}&39.38/0.956 &\underline{38.01}/0.964  &33.67/0.905&32.09/0.901 &27.69/0.777&25.37/0.722 \\
				&&TFPNP \cite{13}&39.72/0.965&37.60/\underline{0.965} &30.49/0.844&28.44/0.783 &27.95/0.795&\underline{26.25}/0.725 \\
                &&DeepMMSE \cite{25}&38.23/0.961&35.95/0.941&32.71/0.885&30.05/0.835&25.83/0.731&23.03/0.636  \\
				&&PrComplex \cite{12}& \underline{40.07}/\underline{0.967}&37.81/0.963 &\underline{33.93}/\underline{0.907}&\underline{32.31}/\underline{0.904} &\underline{27.97}/\underline{0.806}&25.86/\underline{0.761}  \\
				&&PRISTA-Net (Ours)&\textbf{40.55}/\textbf{0.972}&\textbf{38.78}/\textbf{0.970} &\textbf{34.69}/\textbf{0.921}&\textbf{32.72}/\textbf{0.909} &\textbf{29.85}/\textbf{0.825}&\textbf{27.39}/\textbf{0.779} \\
				\\[-5pt]				
			  	&\multirow{6}{*}{$\bm{J=3}$}
                &HIO \cite{2}&34.64/0.870&31.42/0.920 &25.43/0.527&24.19/0.655 &17.00/0.217&13.82/0.272 \\
			  	&&PrDeep \cite{9}&40.18/0.959&39.08/0.969  &34.90/\underline{0.923}&33.39/0.916 &28.64/0.810&26.55/0.779 \\
			  	&&TFPNP \cite{13}&37.65/0.899&\underline{39.20}/\underline{0.971} &31.50/0.868&29.12/0.803 &28.19/\underline{0.814}&\underline{27.05}/\underline{0.781} \\
                &&DeepMMSE \cite{25}&39.79/0.972&37.85/0.960&33.85/0.908&31.48/0.870&26.86/0.775&24.45/0.690  \\
			  	&&PrComplex \cite{12}& \underline{41.26}/\underline{0.973}&39.10/0.971 &\underline{34.93}/0.920&\underline{33.43}/\textbf{0.919} &\underline{28.80}/\textbf{0.833}&26.75/\textbf{0.793}  \\
			  	&&PRISTA-Net (Ours)&\textbf{41.53}/\textbf{0.976}&\textbf{39.94}/\textbf{0.976} &\textbf{35.03}/\textbf{0.932}&\textbf{33.50}/\underline{0.918} &\textbf{29.49}/0.779&\textbf{27.26}/0.728 \\
				\\[-5pt]
				&\multirow{6}{*}{$\bm{J=4}$}
                &HIO \cite{2}&36.08/0.903&31.43/0.940 &26.72/0.585&26.01/0.715 &18.18/0.262&15.23/0.330 \\
				&&PrDeep \cite{9}&40.78/0.961 &39.87/0.973  &35.60/\underline{0.933}&\underline{34.23}/\underline{0.932} &\underline{29.25}/0.833&27.25/0.815 \\
				&&TFPNP \cite{13}&40.59/0.956&\underline{40.05}/0.972 &30.88/0.857&29.66/0.816 &27.44/0.837&\underline{27.60}/\underline{0.827} \\
                &&DeepMMSE \cite{25}&40.81/\underline{0.978}&39.10/0.969&34.18/0.921&32.51/0.893&27.61/0.805&25.34/0.730  \\
				&&PrComplex \cite{12}& \underline{42.08}/0.977&39.99/\underline{0.975} &\underline{35.63}/0.928&34.16/0.929 &29.17/\underline{0.846}&27.27/0.813  \\
				&&PRISTA-Net (Ours)&\textbf{42.78}/\textbf{0.982}&\textbf{41.18}/\textbf{0.981} &\textbf{36.59}/\textbf{0.945}&\textbf{34.55}/\textbf{0.935} &\textbf{31.49}/\textbf{0.870}&\textbf{29.01}/\textbf{0.830} \\
         \bottomrule[1.5pt]
    \end{tabular*}
\end{table*}

\begin{figure*}[!htbp]
\centering
\includegraphics[width=1.0\linewidth]{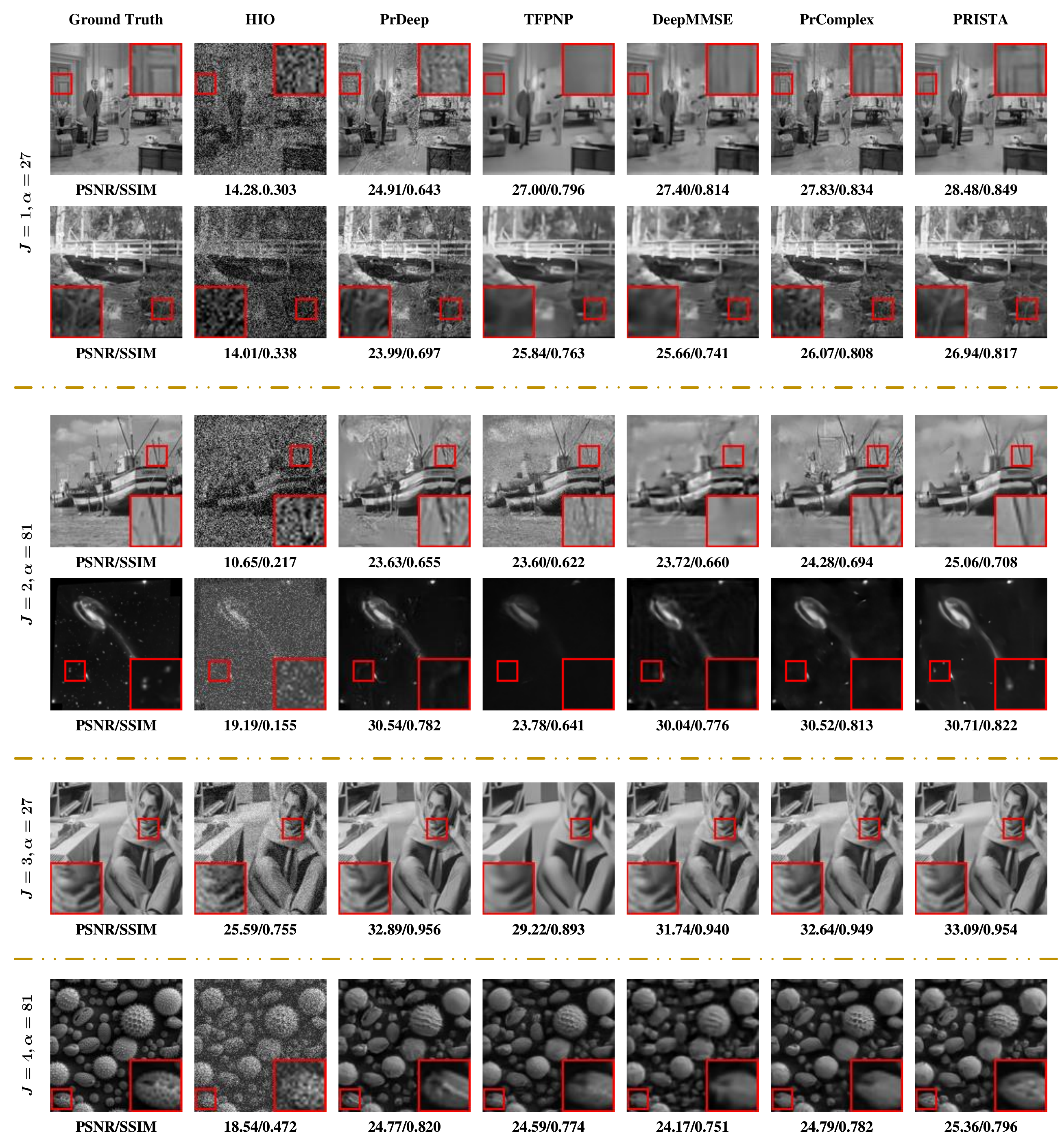}
\caption{Reconstruction comparisons of 128$\times$128 images from noisy intensity-only CDPs measurements using different numbers of uniform masks at different noise levels.}
\label{fig:result_128x128}
\end{figure*}

\subsection{Comparison with State-of-the-Art Methods}
We compare the proposed method with several state-of-the-art algorithms, including HIO\cite{2}, PrDeep\cite{9}, TFPnP\cite{13}, PrComplex\cite{12}, and DeepMMSE\cite{25}. We implement these methods using their publicly available codes and evaluate their performance on the same public test datasets.

\subsubsection{Quantitative evaluation}
Quantitative comparisons of different methods are performed using different numbers of masks at different noise levels on the $128\times128$ and $256\times256$ test datasets. Table \ref{table1} presents the average PSNR and SSIM for each method. The best and second-best results are highlighted in bold and underlined, respectively. We can observe that our method outperforms the other five methods in terms of different numbers of masks and noise levels. For example, when $J=3$, our method yields an average improvement of 0.35 dB, 0.73 dB, and 1.42 dB in UNT-6 at different noise levels. When $J=4$, the average improvements are 0.91 dB, 0.79 dB and 1.64 dB in UNT-6 at different noise levels, indicating that our method is more robust to higher levels of noise.

\begin{figure*}
\centering
\includegraphics[width=1.0\linewidth]{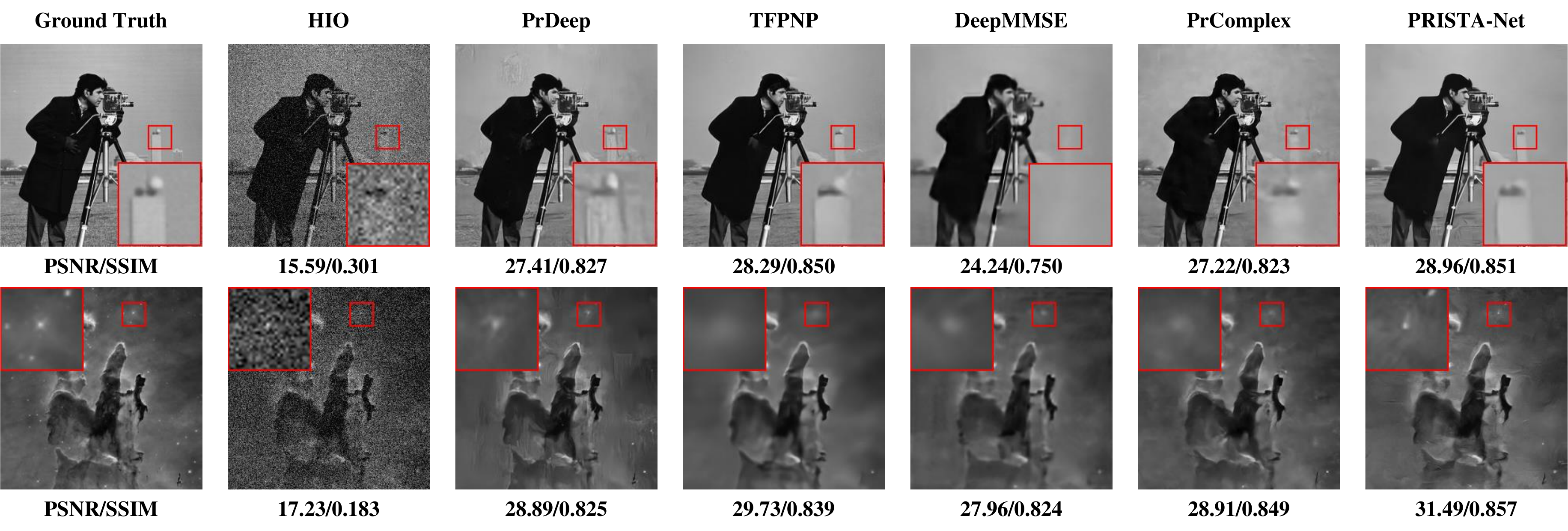}
\caption{Reconstruction comparisons of 256$\times$256 images from noisy intensity-only CDPs measurements using $4$ uniform masks when $\alpha$=81.}
\label{fig:result_256x256}
\end{figure*}

\begin{figure*}
\centering
\includegraphics[width=1.0\linewidth]{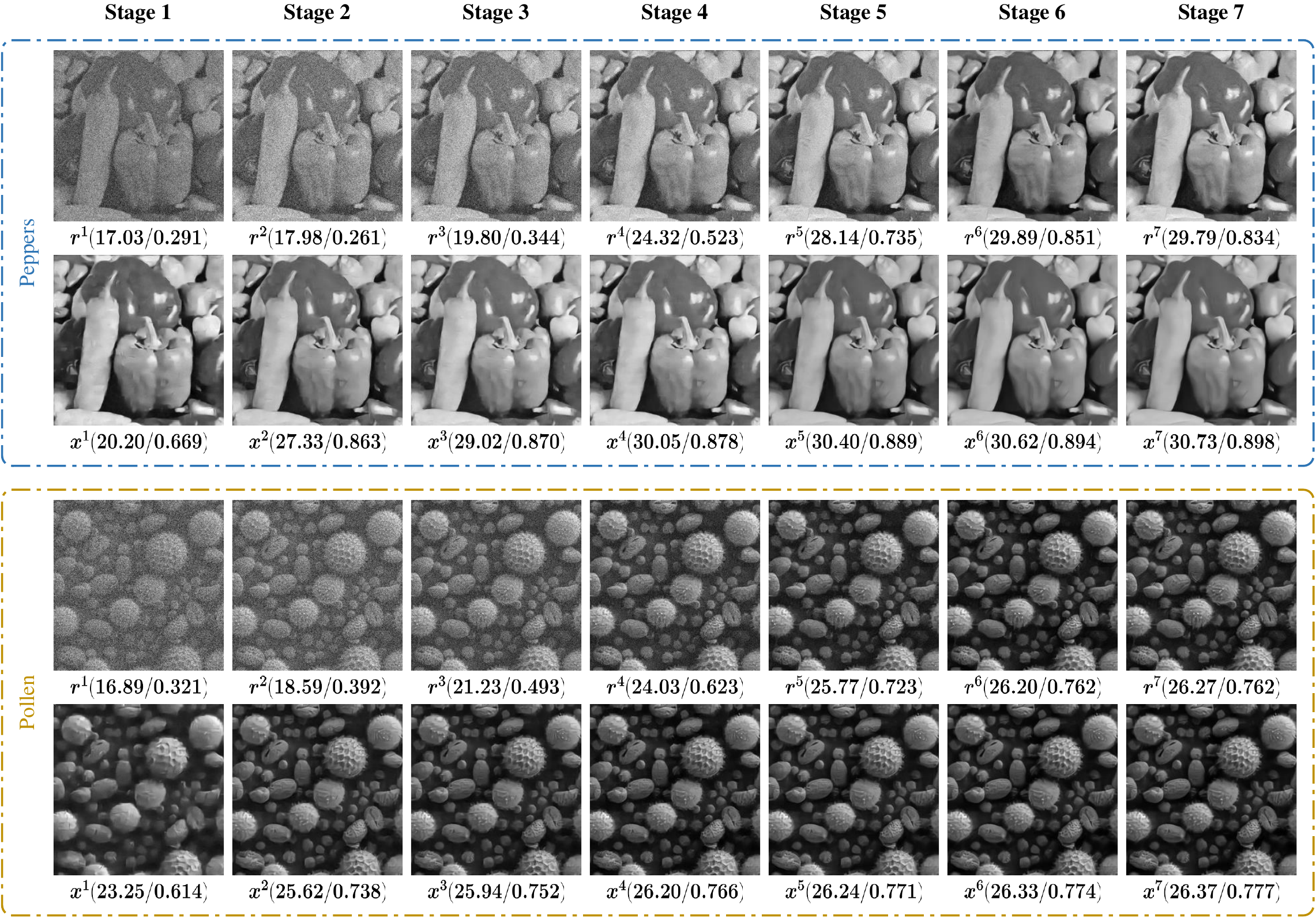}
\caption{The intermediate reconstructions of two images ($256\times256$ Peppers and Pollen) using $4$ uniform masks when $\alpha=81$. }
\label{fig:stageout}
\end{figure*} 
The HIO algorithm is based on the traditional alternating projection method, but has the disadvantage of converging to local minima and being vulnerable to noise. When $\alpha=81$, the average PSNR scores for UNT-6 and NT-6 were even lower than 20 dB. The PrDeep, TFPnP, and PrComplex PnP algorithms, which are based on pre-trained denoisers, have similar reconstruction performance. TFPnP uses reinforcement learning to adjust internal parameters, but the results are not satisfactory because it does not use consistent training masks during testing. In contrast, our method can achieve good reconstruction results even with different masks, which demonstrates its generalizability and makes it suitable for practical applications, such as designing optical modulators. The unsupervised algorithm DeepMMSE does not need real images, but requires retraining the generative network for each image, leading to slow inference due to sample-wise optimization and making it difficult to apply in practice. Our approach, on the other hand, utilizes a well-trained model and benefits from GPU acceleration, resulting in fast inference speed. 

Similarly to the $128\times128$ test dataset, the proposed method generally outperforms existing methods in UNT-6 and NT-6 with the size $256\times256$. When $J=4$, our method achieves an average improvement of 0.7 dB, 0.96 dB, and 2.24 dB in UNT-6 at different noise levels.

In addition, we observe that the reconstruction results for natural images are slightly lower than those for unnatural images, which is likely because natural images contain more detail. As the number of masks increases, the number of measurements also increases, resulting in more information about the amplitude. This improvement in the reconstruction effect is in line with what we would expect.

\subsubsection{Qualitative evaluation}
Fig. \ref{fig:result_128x128} shows the visualization comparisons using different numbers of masks at different noise levels on the $128\times128$ test dataset. Our method can still obtain better reconstruction results with fewer artifacts and richer details compared to other algorithms. For example, in the enlarged region of the couple image (the first row), our algorithm can reconstruct the structure and texture of the door more accurately, while the results of other algorithms are not satisfactory. Similarly, in the enlarged area of the boat image (the third row), our algorithm can clearly reconstruct the mast of the boat. Fig. \ref{fig:result_256x256} shows the results of the comparison between four masks in the $256\times256$ test dataset when $\alpha=81$. 

Our method is evidently superior to the other algorithms, which can be attributed to the use of convolutional operations in both spatial and frequency domains, allowing for the learning of both local and global features. Furthermore, the incorporation of CBAM encourages the model to pay more attention to phase information, such as edges, textures, and structures, and the designed logarithmic-based loss function can obtain better results when the noise level is low.

\subsection{ Analysis for Intermediate Results}
We can further demonstrate the effect of PRISTA-Net on low-quality images as the number of stages increases. Fig. \ref{fig:stageout} shows the outputs of each stage of the SGD and PPM modules for the Peppers and Pollen images ($256\times256$) when $\alpha=81$. It is evident that the PPM module works as a denoising process, while the SGD module helps to sharpen the details of the image, bringing them closer to the ground truth. As the stage continues, artifacts are eliminated and details of the underlying region of interest are improved, resulting in a high-quality image with reduced noise and enhanced textures, and the designed logarithmic-based loss function can obtain better results when the noise level is low.

\section{Conclusions}
In this study, we integrate the ISTA algorithm for the PR problem by truncating the two-step iterative process and replacing the nonlinear transformations in the proximal-point mapping subproblem with a well-designed network, and propose a novel DUN dubbed PRISTA-Net. This method combines the interpretability of traditional methods with the powerful learning ability of deep learning, giving us a new perspective to design the explainable network. In addition, we utilize convolutions in both spatial and frequency domains to capture local and global information, and incorporate attention mechanisms to focus on phase information containing image edges, textures, and structures, thus improving PR. Furthermore, the designed logarithmic-based loss yields considerable enhancements when the noise level is low. Extensive experiments on CDP measurements show that our approach outperforms existing state-of-the-art algorithms while maintaining fast inference speed.
\bibliographystyle{IEEEtran} 
\bibliography{PRISTA_ref}
\end{document}